\documentclass[10pt,twocolumn,letterpaper]{article}

\usepackage{cvpr}              

\usepackage[accsupp]{axessibility}

\usepackage{url}

\usepackage{booktabs}       
\usepackage{amsfonts}       
\usepackage{nicefrac}       
\usepackage{microtype}      

\usepackage{multirow}
\usepackage{balance}
\usepackage{makecell}
\usepackage{algorithm,algorithmic}
\usepackage{pifont}       
\usepackage{wrapfig}
\usepackage{amsmath} 
\usepackage[dvipsnames,table]{xcolor}
\usepackage{graphicx}

\definecolor{Gray}{gray}{0.9}
\definecolor{cvprblue}{rgb}{0.21,0.49,0.74}  
\usepackage[pagebackref,breaklinks,colorlinks,allcolors=cvprblue]{hyperref}

\def\paperID{13490} 
\def\confName{CVPR}
\def\confYear{2025}

\title{FedAWA: Adaptive Optimization of Aggregation Weights in \\Federated Learning Using Client Vectors}

\author{Changlong Shi$^1$, He Zhao$^2$, Bingjie Zhang$^1$, Mingyuan Zhou$^3$, Dandan Guo$^{1*}$, Yi Chang$^{1}$\thanks{ Corresponding authors.} \\
Jilin University, China$^1$ CSIRO’s Data61, Australia$^2$\\ The University of Texas at Austin, USA$^3$\\
\texttt{\{shicl22,zhangbj24\}@mails.jlu.edu.cn,he.zhao@data61.csiro.au}\\
\texttt{mingyuan.zhou@mccombs.utexas.edu,
\{guodandan,yichang\}@jlu.edu.cn} 
}

\begin{document}
\maketitle
\begin{abstract}

Federated Learning (FL) has emerged as a promising framework for distributed machine learning, enabling collaborative model training without sharing local data, thereby preserving privacy and enhancing security. However, data heterogeneity resulting from differences across user behaviors, preferences, and device characteristics poses a significant challenge for federated learning. Most previous works overlook the adjustment of aggregation weights, relying solely on dataset size for weight assignment, which often leads to unstable convergence and reduced model performance. Recently, several studies have sought to refine aggregation strategies by incorporating dataset characteristics and model alignment. However, adaptively adjusting aggregation weights while ensuring data security—without requiring additional proxy data—remains a significant challenge. In this work, we propose Federated learning with Adaptive Weight Aggregation (FedAWA), a novel method that adaptively adjusts aggregation weights based on client vectors during the learning process. The client vector captures the direction of model updates, reflecting local data variations, and is used to optimize the aggregation weight without requiring additional datasets or violating privacy. By assigning higher aggregation weights to local models whose updates align closely with the global optimization direction, FedAWA enhances the stability and generalization of the global model. Extensive experiments under diverse scenarios demonstrate the superiority of our method, providing a promising solution to the challenges of data heterogeneity in federated learning.

\end{abstract}

\section{Introduction}

Federated learning (FL) as an innovative paradigm in machine learning, has garnered significant attention in recent years \citep{survey3, survey1, survey4}. This distributed optimization method leverages multiple local clients to collaboratively train a shared global model without sharing the client data, thereby preserving privacy and enhancing security. FL has a wide range of applications in many areas, such as healthcare \citep{healthy}, finance \citep{financial}, and education \citep{education}. 
The FedAvg \cite{fedavg} algorithm, a cornerstone method in FL, stores a global model on a central server. During the training process, this global model is distributed to participating clients for local updates. After each client updates its model, the server collects and aggregates the optimized parameters from the clients to update the global model. In FedAvg, the aggregation weights are determined based on the size of the local datasets. However, the performance of the model optimized by FedAvg tends to degrade due to data heterogeneity, which arises from user behaviors, preferences, devices, organizations, and other diverse factors \citep{survey2}. As a result, different local models tend to optimize towards distinct local objectives, causing divergent optimization directions and unstable convergence \citep{fedprox}, and ultimately degrading the overall model performance. This phenomenon has been both theoretically and empirically validated in \citep{non-iid2, non-iid3}.
\vspace{-0.5em}

To mitigate the negative effects of data heterogeneity, several approaches have been proposed that adjust the aggregation weights on the server side to reduce bias during the model aggregation process.
FedDisco \citep{feddisco} leverages both dataset size and the discrepancy between local and global category distributions to determine more distinguishing aggregation weights. Similar to FedAvg, FedDisco also uses fixed aggregation weights throughout the training process, which limits its ability to adapt to the dynamic optimization process. L-DAWA \citep{ldawa} employs cosine similarity between local models and the global model as aggregation weights, preventing deviations in the aggregation process from the global optimization direction. However, the similarity between models may not fully capture the relationship between clients, and directly using it as aggregation weights may lack sufficient adaptability.
FedLAW \citep{fedlaw} identifies the global weight shrinking phenomenon and learns the optimal aggregation weights at the server with a proxy dataset, which is assumed to have the same distribution as the global dataset. This raises privacy concerns, which are of paramount importance in federated learning. Therefore, adaptively adjusting aggregation weights while ensuring data security remains a significant challenge.

\vspace{-0.2em}
To this end, we draw inspiration from recent work on task arithmetic \citep{task-arithmetic, tiemerging}. It suggests that changes in model parameters during training (referred to as task vectors) often capture meaningful information about the datasets and can be directly manipulated through arithmetic operations. Since direct access to clients' local data is not feasible in the federated learning context, we hope to leverage a similar approach about task arithmetic, $i.e.,$ the changes in model parameters after local training, to infer information about the clients' local data. Given that the heterogeneity in federated learning primarily stems from differences in clients' local data, in this paper, we replace the original concept of the \textit{task vector} with \textit{client vector}, which is derived by subtracting the global model parameters from the locally trained client model parameters. 
We then conduct experiments to investigate the relationship between the client vector and local data, empirically demonstrating that client vectors capture the variations among local datasets. Furthermore, the aggregated client vectors align more closely with the desired direction of model updates. These findings could serve as a valuable guide for improving the model aggregation process in federated learning.

\vspace{-0.2em}
Building on these observations, we introduce Federated learning with Adaptive Weight Aggregation (FedAWA), a method that requires no additional datasets and enables dynamic adjustment of aggregation weights throughout the training process. FedAWA optimizes the aggregation weights on the server side, assigning higher weights to local models whose client vectors are more aligned with the overall update direction, thereby enhancing both the global model generalization and the training stability. To validate the effectiveness of FedAWA, we conduct extensive experiments across diverse scenarios.
The contributions of this work are summarized  as follows: 
\begin{itemize}
\vspace{0.75em}
\item[$\bullet$] We investigate the relationship between model merging and federated learning, designing the client vector to optimize aggregation weights in federated learning.
\vspace{0.75em}
\item[$\bullet$] We introduce FedAWA, a simple yet effective method for adaptively adjusting aggregation weights on the server side. FedAWA requires no additional data or transmission of original data, thus raising no privacy concern.
\vspace{0.75em}
\item[$\bullet$] We conduct extensive experiments across diverse scenarios, demonstrating the effectiveness of FedAWA.
\vspace{0.75cm}

\end{itemize}

\section{Related Works}
\label{sec:relatedwork}

\textbf{Federated Learning.}
Federated learning is a rapidly advancing research field with many remaining open problems to address.
One of the primary issues that can significantly degrade its performance is data distribution heterogeneity. Research addressing this problem can generally be categorized into two main directions: client-side and server-side adjustment.
First is the client-side drift adjustment.
Due to the data heterogeneity, local models trained on the clients may exhibit different degrees of bias, thereby affecting the performance of the global model. 
Many methods aim to reduce this bias by adjusting the training process of local models \cite{feddc, fedetf, moon}.
FedProx \citep{fedprox} utilizes the $l_2$-distance between the global model and the local model as a regularization term during the training of the local model. FedDyn \citep{feddyn} proposes a dynamic regularizer for each client to align the global and local solutions. 
These methods merely assign aggregation weights based on the size of the local dataset. In contrast, our approach focuses on the server-side aggregation process, making it easily combined with these methods to enhance model performance further.
Several other studies focus on adjusting the model on the server side \cite{fedbe, feddf}.
FedDisco \citep{feddisco} leverages both dataset size and the discrepancy between local and global category distributions to design the fixed aggregation weights.
FedLAW \citep{fedlaw} revisits the weighted aggregation process and utilizes a proxy dataset, which is assumed to have the same distribution as the global dataset, to learn the optimal global weight shrinking factor and the aggregation weights. 
In contrast, our proposed FedAWA eliminates the need for proxy datasets, while also enables the dynamic optimization of aggregation weights.
L-DAWA \citep{ldawa} employs cosine similarity between local models and the global model as aggregation weights. Both L-DAWA and ours can adaptively adjust the aggregation weights during the training process.
The key difference is that we optimize the aggregation weights rather than directly using the similarity as the aggregation weight. Besides, we leverage the variations of local model and global model before and after training to design the client vector and global vector, which better reflects the local data information.

\textbf{Model Merging.}
Recently, with the rapid advancements in deep learning, model merging has garnered significant attention \citep{modelsoup, fisher_merging, git_re}. This technique aims to aggregate multiple well-trained models into a single unified model, thereby inheriting their individual capabilities without incurring the computational overhead and complexity associated with traditional ensembling methods. Task Arithmetic \citep{task-arithmetic} stands out as a simple yet highly effective method for model merging. It introduces task vectors, which are both efficient and lightweight, facilitating improved generalization across tasks. Ties-Merging \cite{tiemerging} identifies redundancy and sign conflicts in direct task vector aggregation and proposes three steps to resolve them. AdaMerging \cite{adamerging} addresses the limitation of shared merging coefficients in prior methods, designing separate aggregation weights for each task vector to enhance model adaptability. The ability of model merging to aggregate models without relying on training data closely aligns with the requirements of federated learning, where preserving data privacy is of paramount importance. However, model merging is a single-step aggregation and the federated learning needs a multiple communication rounds, where the former already owns well-trained local models and the latter requires constant iterative updating of the global and local models. The key difference makes it difficult to design the aggregation weights in federated learning using the model merging directly, which is ours research focus in this work. To the best of our knowledge, we are the first to explore the relationship between model merging and federated learning, and introduce an adaptive optimization of aggregation weights by designing the client vector in federated learning.

\vspace{-0.5em}
\section{Background}
\label{background}
\textbf{Federated Learning.}
Federated Learning consists of $K$ clients and a central server, where each client has its own private local dataset $\mathcal{D}_k$. FL aims to enable clients to collaboratively learn a global model for the server without data sharing. 
In communication round $t$ (out of a total of $T$ rounds), the parameters of the global model and the client $k$'s model are denoted as $\theta^t_g$ and $\theta^t_k$, respectively. The workflow of the basic FL method, FedAvg \citep{fedavg}, in communication round $t$ can be described as follows:
\begin{itemize}
\vspace{0.2em}
\item \textbf{Step 1:} Server broadcasts the parameters of global model $\theta^t_g$ to each client;
\vspace{0.2em}
\item \textbf{Step 2:} Each client $k$ performs $E$ epochs of local model training on private dataset $\mathcal{D}_k$ to obtain a local model $\theta^t_k$;
\vspace{0.2em}
\item \textbf{Step 3:} Clients upload the local models to the server;
\vspace{0.2em}
\item \textbf{Step 4:} Server merges the local models to get a new global model: $\theta^{t+1}_g=\sum_{k=1}^K \lambda_k \theta^t_k$, where $\lambda_k$ is the aggregation weight of the client $k$ and FedAvg sets $\lambda_k=\frac{|\mathcal{D}_k|}{\sum_{i=1}^K|\mathcal{D}_i|}$.
\vspace{0.2em}
\end{itemize}

However, simply using dataset size as aggregation weights in step 4 is suboptimal when local data exhibits high heterogeneity \cite{non-iid1,non-iid2,non-iid3}. 
Various methods have attempted to address this issue by adjusting the aggregation scheme \citep{feddisco,ldawa, fedlaw}, but they still face challenges such as lack of adaptability and the requirement for the proxy dataset. 
In this paper, we aim to explore a method that can adaptively adjust the model aggregation weights ($\boldsymbol{\lambda}$ in step 4) during the training process, while eliminating the need for a proxy dataset. This method is designed to enhance both the performance of the global model and privacy security.

\textbf{Task Arithmetic.} 
Task Arithmetic \citep{task-arithmetic} stands out as a straightforward yet highly efficient technique for merging models. It introduces task vector, which is defined as the difference between the fine-tuned and pre-trained model parameters, i.e., $\tau_k = \theta_k^* - \theta_0$, where $\theta_0$ represents the pre-trained model parameters, and $\theta_k^*$ refers to the model parameters fine-tuned on the downstream task $k$. 
The core idea of this approach involves summing task vectors, which are scaled and added to the pre-trained model parameters to compute the final parameters of the merged model. This can be mathematically formulated as $\theta' = \theta_0 + \lambda \sum_{i=1}^{n} \tau_i,$
where $\lambda$ is the scaling coefficient, and $\theta'$ represents the aggregated model parameters.
Task arithmetic's advantage lies in its ability to integrate model parameters without needing access to the original training data, while producing a merged model that performs well across tasks. This capability closely aligns with the requirements of federated learning, where preserving data privacy is of paramount importance. However, task arithmetic is typically characterized by a single-step model aggregation, and federated learning involves multiple communication rounds with iterative local model training on the client side and model aggregation on the server side, which is the major challenge in federated learning. 



\begin{figure}[t]

  \centering
  
   \includegraphics[width=0.45\textwidth]{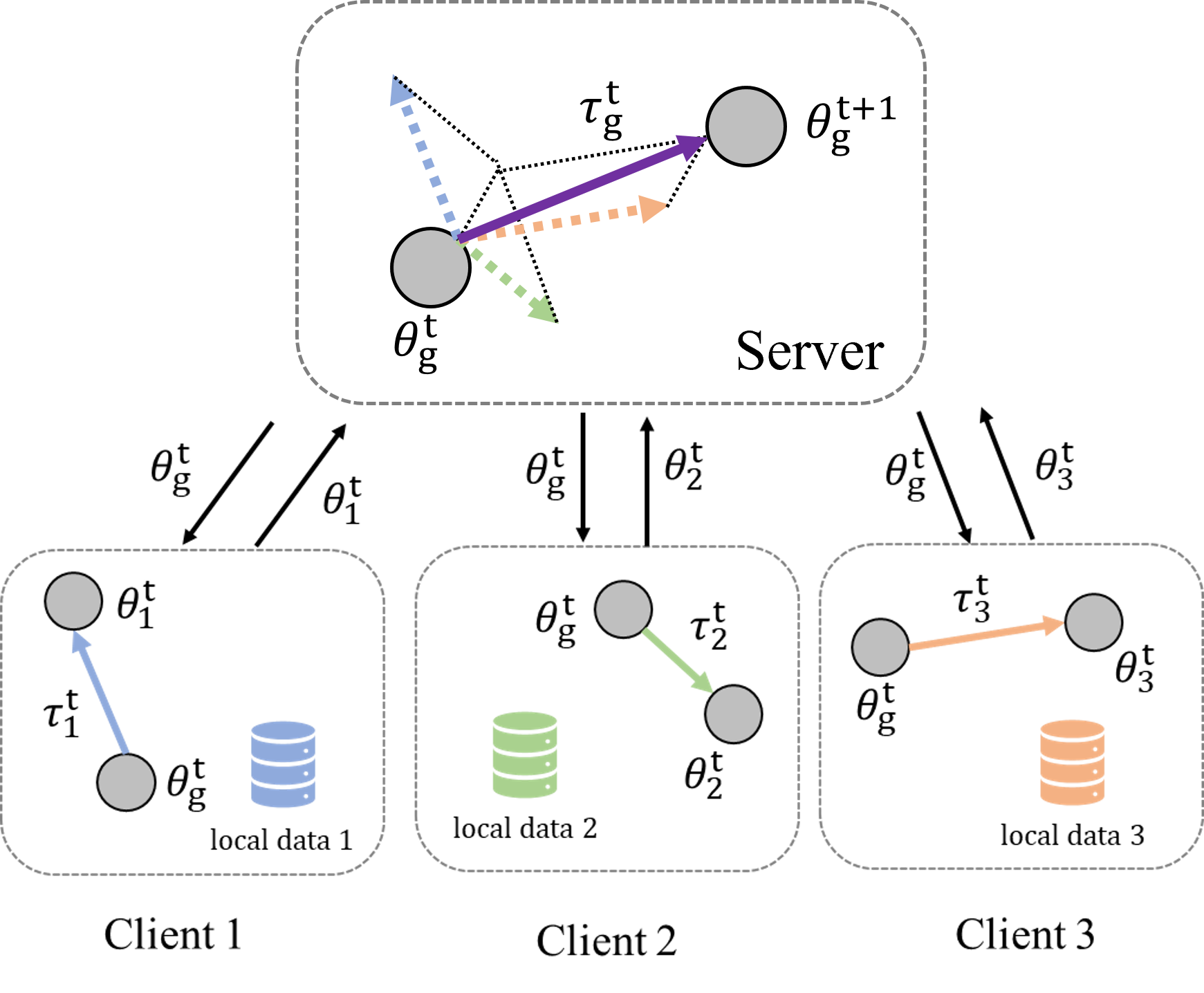}

   \caption{The illustration of client vectors. $\tau^t_k$ is the client vector of the $k$-th client, and $\tau^t_g$ represents the global vector obtained by aggregating the client vectors.}
   \label{client_vector}
   \vspace{-1em}
\end{figure}

\vspace{-0.5em}
\section{Method}

In this section, we introduce Federated Learning with Adaptive Weight Aggregation (FedAWA), a method that adaptively optimizes the aggregation weights without relying on a proxy dataset, thereby enhancing model performance while effectively addressing privacy concerns. Below, we introduce the details of our proposal.

\begin{figure*}[t]
  \centering
  \begin{subfigure}[b]{0.25\textwidth}
    \centering
    \includegraphics[width=\textwidth]{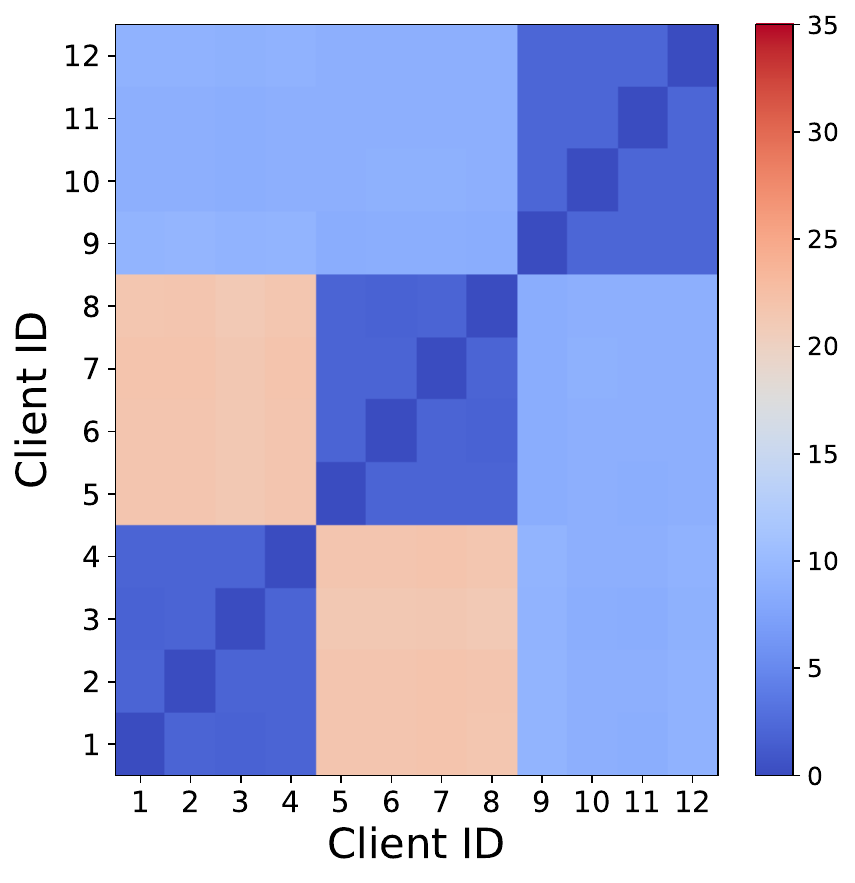}
    \caption{Difference between client datasets.}
    \label{matrix_data}
  \end{subfigure}
  \hfill
  \begin{subfigure}[b]{0.25\textwidth}
    \centering
    \includegraphics[width=\textwidth]{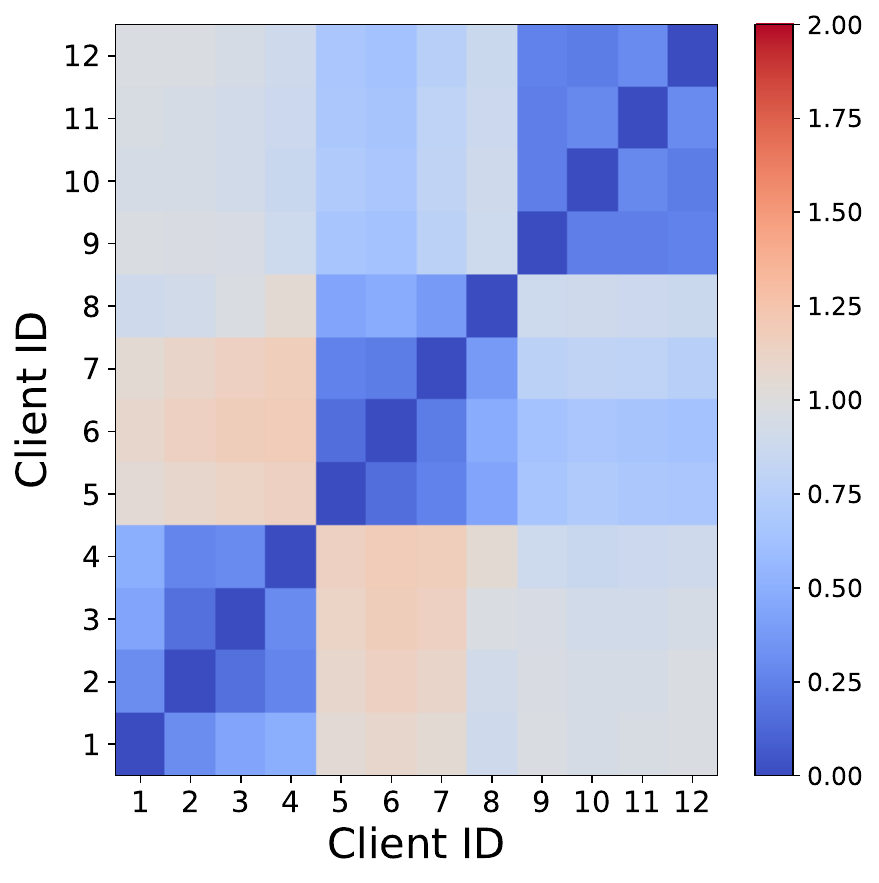}
    \caption{Difference between client vectors.}
    \label{matrix_client_vector}
  \end{subfigure}
  \hfill
  \begin{subfigure}[b]{0.25\textwidth}
    \centering
    \includegraphics[width=\textwidth]{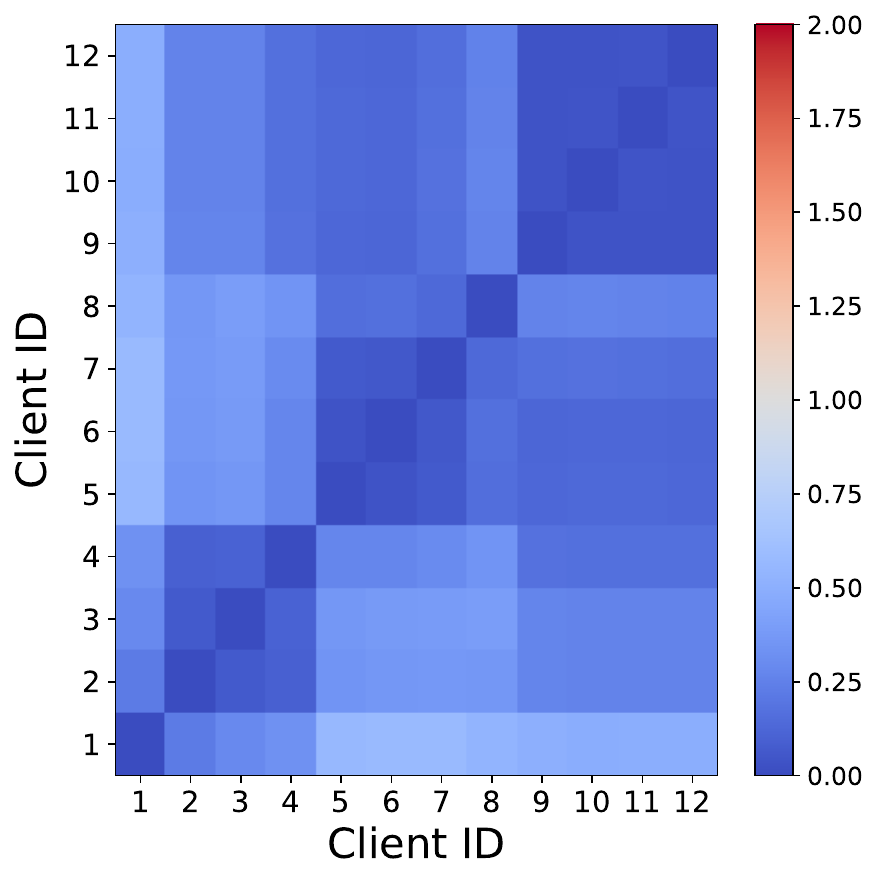}
    \caption{Difference between client models.}
    \label{matrix_model}
  \end{subfigure}
    
  \caption{Differences between client local datasets $\mathcal{D}_{1:K}$, client vectors $\tau_{1:K}^t$, and client models $\theta_{1:K}^t$.}
  \vspace{-1em}
  \label{fig:distance}
\end{figure*}

\vspace{-0.5em}
\subsection{Motivation}
The motivation for our work derives from the task arithmetic in the field of model merging that can reflect the task-related information without data leakage.
In the context of federated learning, the server distributes the global model to clients for training at each communication round. The global model serves as the ``per-trained'' model, while training different  client models based on the respective heterogeneous local data can be considered as different ``tasks''. 
Given that federated learning heterogeneity mainly arises from differences in client local data, we replace the \textit{task vector} with the \textit{client vector}, defined as the difference between the global model parameters and the locally trained client model parameters. A more intuitive illustration is shown in Figure \ref{client_vector}.
In the $t$-th communication round, the client vector of the $k$-th client $\tau_k^t$ and the merged global vector $\tau_g^t$ are defined as follows:
\begin{equation}
\begin{aligned}
\label{client_vect_Eq}
\tau_k^t &= \theta^t_k - \theta^t_g, \quad
\tau^t_g &= \sum_{k=1}^K {\lambda_k^t} \tau^t_k,
\end{aligned}
\end{equation}
where $\lambda_k^t$ is the learnable aggregation weight at the $t$-th communication round. Building on previous research \citep{adamerging, task-arithmetic}, we hypothesize that the client vector can more efficiently encapsulate relevant information about the local data. This information can then be leveraged to optimize the aggregation weights in federated learning, all while preserving data privacy. Then, we empirically validate this hypothesis through experiments.

\vspace{-0.5em}
\subsection{Empirical Observations}\label{observation}

\textbf{Client Vector and Local Data.}
We first investigate the relationship between the client vectors and the local data to demonstrate that the client vectors effectively capture the variations between different local data and reflect the corresponding differences among the local datasets.
In Figure \ref{matrix_client_vector}, we illustrate the divergence between different local datasets and between the corresponding client vectors. As shown, the difference between client vectors in Figure \ref{matrix_client_vector} closely resemble those of the local dataset in Figure \ref{matrix_data}. 
The distance between models parameters, as shown in Figure \ref{matrix_model}, fails to effectively reflect the relationships between local datasets. This demonstrates that , compared to previous methods \cite{ldawa} relying on overall model parameters, the client vector provides a more accurate representation of the information in local datasets. Hence, we explore the possibility of leveraging this phenomenon to enhance the model aggregation process in federated learning. More implementation details can be found in the Appendix \ref{app_client_vector}. 

\begin{figure}[t]
  \centering
   \includegraphics[width=0.42\textwidth]{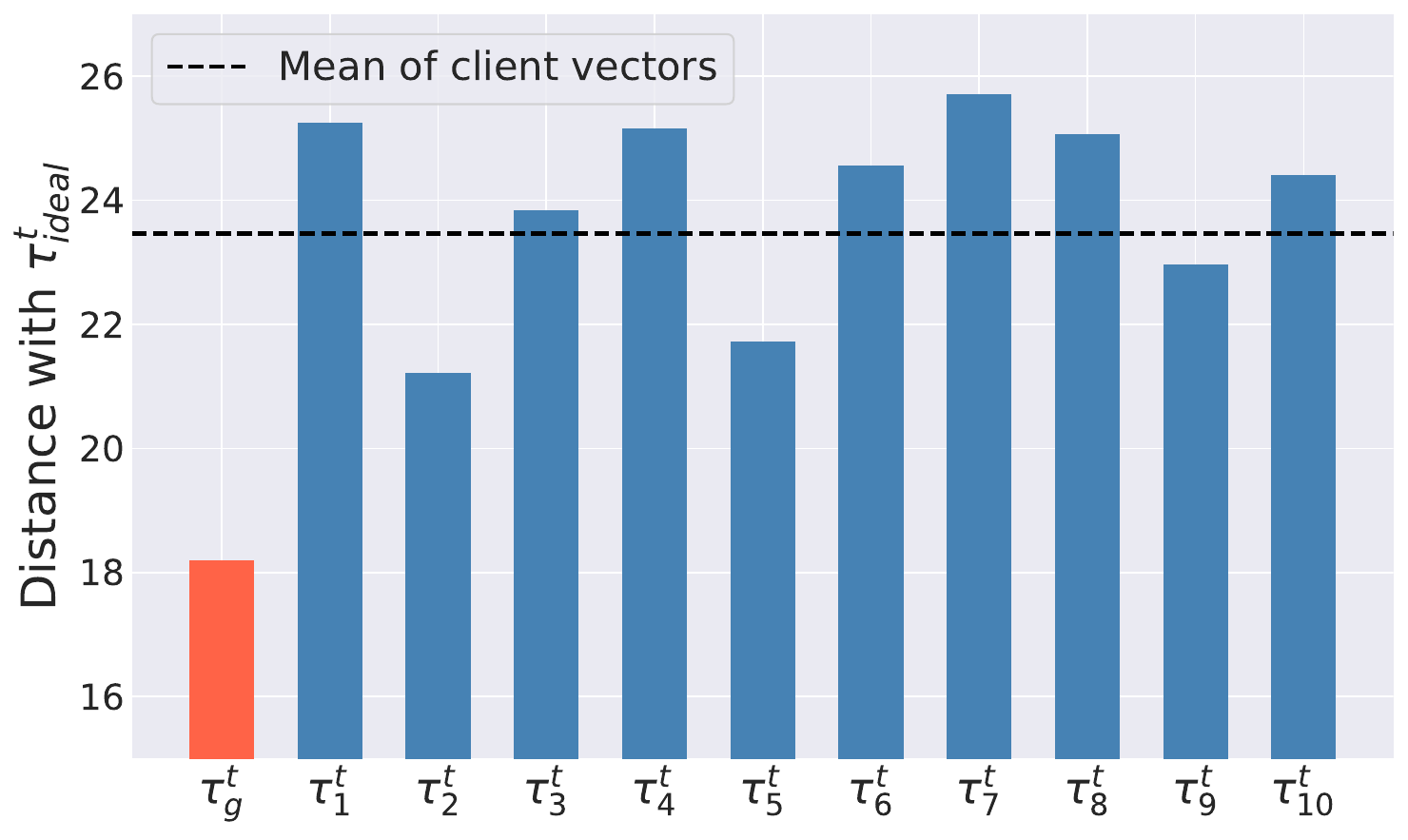}
    \vspace{-1em}
   \caption{Distance with the ideal update vector $\tau^t_{ideal}$.
  }
  \vspace{-1em}
    \label{ideal_vect_distance}
\end{figure}

\textbf{Ideal Vector.} Here, we explore the relationship between the merged global vector and the ideal vector, where the former is computed with the size of the local datasets being the aggregation weight in \eqref{client_vect_Eq} and the latter is explained below. Ideally, federated learning aims to aggregate local client models into a global model that matches the performance of a model trained directly on the global dataset $\mathcal{D}_{g}=\mathcal{D}_1 \cup \mathcal{D}_2 \cup ... \cup \mathcal{D}_K $, where $\mathcal{D}_k$ is the local dataset of the $k$-th client. Now, we view $\theta^t_g$ in \eqref{client_vect_Eq} as the initialized global parameter and use the global dataset $\mathcal{D}_{g}$ to train the model, producing the  parameter $\theta^{t}_{ideal}$. Notably, in practical federated learning, clients do not share data with each other, and we only utilize the global dataset $\mathcal{D}_{g}$ to explore potential methods for enhancing the effectiveness of federated learning.  Now, we can denote $\tau^t_{ideal}$ as the ideal vector in the $t$-th communication round, expressed as $\tau^t_{ideal}=\theta^{t}_{ideal}-\theta^t_g$. To explore whether aggregating the client vectors could similarly capture the update direction of the model trained on the global dataset, we compare the distance between merged global vector (and all client vectors) and the ideal vector $\tau^t_{ideal}$. As shown in Figure \ref{ideal_vect_distance}, it can be observed that the aggregation of the client vectors $\tau^t_g$ is closer to the ideal vector $\tau^t_{ideal}$ compared to individual client vectors $\tau^t_{k}$. This indicates that aggregating the client vectors can yield a pseudo global vector that reflects the overall data distribution across all clients. Therefore, we will utilize the client vector and the merged global vector as the guide to optimize the aggregation weights in federated learning.

\begin{figure}[t]
  \centering
   \includegraphics[width=0.45\textwidth]{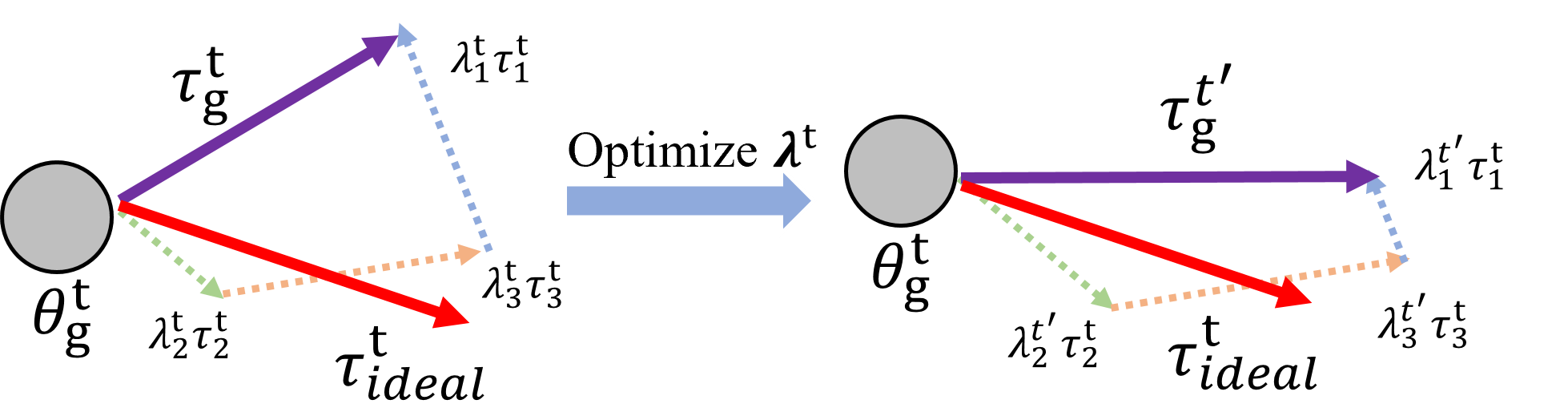}

   \caption{Adjustment of the aggregation weights.
  }
    \label{alin}
\end{figure}

\begin{algorithm}[t]
   \caption{FedAWA: Federated learning with Adaptive Weight Aggregation.}
   \label{alg 1}
\begin{algorithmic}[1]
   \STATE {\bfseries Input:} Communication round $T$, local epoch $E$, local datasets \{$\mathcal{D}_1,...,\mathcal{D}_K $\}, initial global model $\theta^1_{g}$, initial aggregation weights $\boldsymbol{\lambda}^0$, where $\lambda^0_k=\frac{|\mathcal{D}_k|}{\sum_{i=1}^K|\mathcal{D}_i|}$;
   \STATE {\bfseries Output:} Final global model $\theta^{T}_{g}$;
   \FOR{$t=1$ {\bfseries to} $T$}
   \STATE Server sends global model $\theta^{t}_{g}$ to each client;\\
   \renewcommand{\algorithmiccomment}[1]{\# #1}
   \COMMENT{Clients execute:}
   \FOR{each client $k \in [K]$}
   \STATE$\theta^{t}_{k}\gets$ ClientUpdate$(\theta^{t}_g, \mathcal{D}_k,E)$;
   \STATE Send $\theta^{t}_k$ to server;
   \ENDFOR \\
   \COMMENT{Server executes:}
   
    \STATE Server computes $\tau_k^t$, $\tau_g^t$ through Equation \ref{client_vect_Eq}.
    \STATE \textbf{If FedAWA then}: Server optimizes the aggregation weights $\boldsymbol{\lambda^t}$ according to Equation \ref{loss}.
    \STATE \textbf{If FedAWA-L then}: Server optimizes the aggregation weights $\boldsymbol{\lambda^t_l}$ according to Equation \ref{layerwise_loss}.
    \STATE Server aggregates the local models to generate the global model:
    \STATE \textbf{If FedAWA then}: $ {\theta}^{t+1}_g=\sum ^K _{k=1} \lambda_k^{t}\theta^{t}_k$;
    \STATE \textbf{If FedAWA-L then}: $ {\theta}^{t+1}_{gl}=\sum ^K _{k=1} \lambda_{kl}^{t}\theta^{t}_{kl}$;
    
   \ENDFOR
\end{algorithmic}

\end{algorithm}

\vspace{-0.5em}
\subsection{FedAWA}\label{method}

Based on the aforementioned empirical observations, we suggest that in FL, the aggregation of client vectors can serve as an indicator to adjust the aggregation weights. Specifically, clients whose update directions closely align with the merged global vector should be assigned greater weights, reflecting their contributions that are more consistent with the overall learning objective. Conversely, clients with less aligned update directions can be assigned lower weights to minimize potential negative impacts on model convergence. An intuitive example is shown in Figure \ref{alin}. By adjusting the aggregation weights, the deviation between the aggregated global vector and the ideal vector can be minimized, effectively reducing the variation among clients. Thus, we propose FedAWA by leveraging the relationship between client vectors and merged global vector to dynamically adjust the model aggregation weights, thereby enhancing the overall effectiveness and robustness of the aggregation process in federated learning. More specifically, the initial three steps of our method align with those outlined in Section \ref{background}. The server broadcasts the global model \( \theta_g^t \) to each client, and the client sends the locally trained model \( \theta_k^t \) back to the server. Then, we calculate the client vector of communication round $t$ as $\tau_k^t = \theta_k^t - \theta_g^t$. We aim to optimize the aggregation weights by assigning higher weights to clients whose update directions are more alignment to the merged model vector \( \tau_g^t \). Subsequently, the distance between the client vectors and the derived merged model vector \( \tau_g^t \) in \eqref{client_vector} be used as a supervisory signal to optimize the aggregation weights, and the objective function can be expressed as:

\vspace{-1em}

\begin{table*}[t]
  \begin{center}
    \caption{Top-1 test accuracy (\%) on CIFAR-10, CIFAR-100, and TinyImageNet datasets with $\alpha=0.5$, $\alpha=0.1$, and $\alpha=100$.}
    \vspace{-1em}
    \label{main_res}
    \resizebox{\linewidth}{!}{
    \renewcommand{\arraystretch}{1}
    \begin{tabular}{l c c c c c c c c c c}
    \toprule
       \multicolumn{1}{l}{\textbf{Dataset}} & \multicolumn{3}{c}{CIFAR-10} & \multicolumn{3}{c}{CIFAR-100} & \multicolumn{3}{c}{TinyImageNet} & \multirow{2}{*}{\textbf{Average}} \\
       \cmidrule(lr){2-4} \cmidrule(lr){5-7} \cmidrule(lr){8-10}
       \multicolumn{1}{l}{\textbf{Heterogeneity}} & NIID($\alpha$=0.1) & NIID($\alpha$=0.5) & IID($\alpha$=100) & NIID($\alpha$=0.1) & NIID($\alpha$=0.5) & IID($\alpha$=100) &NIID($\alpha$=0.1) & NIID($\alpha$=0.5) & IID($\alpha$=100) \\
      \midrule
      FedLAW \citep{fedlaw} & 64.76 & 75.27 & 81.30 & 34.59 & 37.56 & 41.05 & 29.13 & 33.49 & 37.20 &  48.26\\
       \midrule
        FedAvg \citep{fedavg} & 61.04 & 74.47 &76.01& 36.71 & 41.08 &41.46& 29.44 & 34.43 &36.31& 47.88\\
        FedDisco \citep{feddisco} & 62.86 & 74.72 &75.40& 36.46 & 41.02 & 41.46 & 31.19 & 34.29 & 36.31 & 48.19 \\
        L-DAWA \citep{ldawa} & 62.87 & 75.61 &76.10& 36.31 & 39.81 & 42.39 & 32.02 & 31.43 & 36.25 & 48.09 \\
        FedProx \citep{fedprox} & 60.62 & 73.27 & 73.96 & 34.60 & 39.35 & 38.15 & 29.37 & 34.32 & 35.03 & 46.52\\
        FedAdam \citep{fedadam}&61.76&73.04& 70.40 &32.12&34.92& 30.37 &21.77&27.39& 24.08 &41.76 \\
        FedDyn \citep{feddyn} &56.37&74.61& 77.92 &36.92&\textbf{44.80}& 41.04 &27.32&32.80& 34.32 &47.34 \\
        
        \midrule
        \textbf{FedAWA (Ours)} & 63.55 & 75.65 & \textbf{80.10} & \textbf{37.04} & 41.89 & 42.84 & 33.07 & 34.57 & \textbf{36.59} & 49.48 \\
        \textbf{FedAWA-L (Ours)} & \textbf{65.13} & \textbf{75.99}& 79.70&36.85&42.52&\textbf{45.27}&\textbf{33.42}&\textbf{34.86}&36.04 & \textbf{49.98} \\
        \bottomrule
    \end{tabular}}
    
  \end{center}
  \vspace{-2em}
\end{table*}

\begin{figure*}[t]

  \centering
  
   \includegraphics[width=0.9\textwidth]{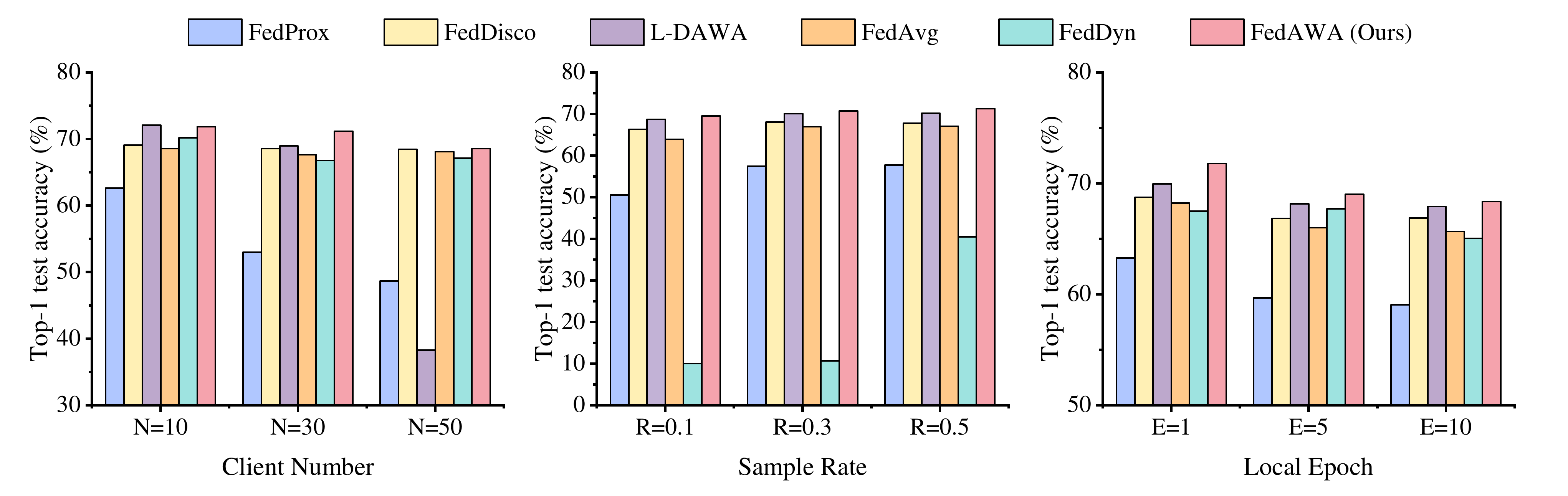}
    \vspace{-0.5em}
   \caption{Top-1 test accuracy (\%) on different client numbers, sample rates, and local epochs.}
   \label{fig_diff_fl_set}
   \vspace{-1em}
\end{figure*}

\begin{equation}
\begin{aligned}
\label{obj1}
{\boldsymbol{\lambda}^{t}} = \arg\min_{\boldsymbol{\lambda}} \left( \sum_{k=1}^{K} \lambda_k \|\tau_k^t-\tau_g^t\|_2 \right), \text{s.t.} \|\boldsymbol{\lambda}\|_1=1.
\end{aligned}
\end{equation}
Through the function, clients whose update directions more align with the merged global vector are assigned higher aggregation weights, while those with less aligned directions receive lower weights. This approach optimizes the global model’s update trajectory, ultimately enhancing overall model performance.
In addition, we also incorporate a global alignment component to ensure that the aggregated global model does not deviate excessively from the previous communication round's global model, thus maintaining the stability of the training process. To achieve this, we introduce an additional regularization term and rewrite the objective function (\ref{obj1}) as follows:
\begin{equation}\label{loss}
 \begin{aligned}
&{\boldsymbol{\lambda}^{t}}=\underset{\boldsymbol{\lambda}}{\arg\min}\left( \sum_{k=1}^{K} \lambda_k \|\tau_k^t\!-\!\tau_g^t\|_2\!+\! \text{d}(\sum_{k=1}^{K} \lambda_k \theta_k^t, \theta_g^{t})  \right),\\ 
& \text{s.t.}
\|\boldsymbol{\lambda}\|_1 =1,
 \end{aligned}
\end{equation}
where d$(\cdot,\cdot)$ represents the distance function. In this paper, we implement it using 1 - cosine similarity. In the experiments, we also evaluate the impact of different distance metrics on model performance.
Following the optimization procedure outlined above, the aggregation weight $\boldsymbol{\lambda}^{t}$ is derived and subsequently utilized in step 4 in Section \ref{background} to produce the final global model $\theta_g^{t+1}=\sum_{k=1}^{K} \lambda_k^{t} \theta_k^t$.

Previous studies have demonstrated significant divergence across various layers of deep neural networks \citep{pfedla, ldawa, fedlama}. Given that each layer in deep neural networks may vary differently, it might be beneficial to design specific aggregation weights for each layer of the model. 
Consequently, we extend our method by calculating separate aggregation weights for each layer and propose FedAWA-L, which allows for more fine-grained adjustments in the model aggregation process.
The objective function of the layer-wise method as:
\begin{equation}\label{layerwise_loss}
 \begin{aligned}
&{\boldsymbol{\lambda}^{t}_l}=\underset{\boldsymbol{\lambda}}{\arg\min}\left( \sum_{k=1}^{K} \lambda_{kl} \|\tau_{kl}^t\!-\!\tau_{gl}^t\|_2\!+\! \text{d}(\sum_{k=1}^{K} \lambda_{kl} \theta_{kl}^t, \theta_{gl}^{t})  \right),\\ 
& \text{s.t.}
\|\boldsymbol{\lambda}\|_1 =1,
 \end{aligned}
\end{equation}
where $\lambda_{kl}$ represents the aggregation weight for the $l$-th layer of the $k$-th client model. $\tau^t_{kl}$ and $\theta_{kl}^t$ represent the $l$-th layer of the client vector and the local model for client $k$, respectively, while $\tau^t_{gl}$ and $\theta_{gl}^t$ denote the $l$-th layer of the global vector and the global model.

Through the aforementioned method, we can adaptively optimize the aggregation weights during the federated learning training process without requiring additional information. All the utilized information comes from the existing information within the basic federated learning algorithm, thereby ensuring privacy security.  The pseudo-code of FedAWA is shown in Algorithm \ref{alg 1}.

\vspace{-0.5em}
\subsection{Discussions}

\textbf{Privacy.}
FedAWA offers enhanced privacy protection and practical adaptability over prior approaches \citep{feddisco, fedlaw}. Instead of relying on fine-tuning with a proxy dataset, it directly optimizes aggregation weights by leveraging model parameters and gradient updates, both of which are easily accessible to the server during training. By eliminating the need for additional data, FedAWA mitigates the risk of data leakage, making it more applicable to real-world environments.

\vspace{0.2em}
\noindent\textbf{Modularity.}
Our proposed FedAWA can serve as a plug-and-play module for many existing FL methods, enhancing their performance across a wide range of applications. For FL methods that adjust the client-side model \citep{fedprox,feddyn,moon}, FedAWA operates on the server-side, making it easily integrable with these client-side adjustments. Furthermore, for methods that adjust the server-side model \citep{fedavg,feddisco,fedavgm}, which use fixed aggregation weights, these weights can be used as initial values in our optimisation process, allowing for further optimisation and improved model performance.


\vspace{-1em}
\section{Experiments} \label{experiment}

\vspace{-0.5em}
\subsection{Experiment Setup}

\textbf{Dataset and Baselines.} In this study, we consider three image classification datasets: CIFAR-10 \citep{cifar}, CIFAR-100 \citep{cifar}, and Tiny-ImageNet \citep{tinyimagenet}. For each dataset, all methods are evaluated with the same model architectures for a fair comparison. In Table \ref{main_res}, We use ResNet20 \citep{resnet} for CIFAR-10 and CIFAR-100, ResNet18 \citep{resnet} for Tiny-ImageNet.
We compare our method with seven representative baselines. Specifically, (1) FedAvg \citep{fedavg} serves as the standard algorithm for Federated Learning; (2) FedProx \citep{fedprox}, and FedDyn \citep{feddyn} focus on local model adjustments; (3) FedAdam \cite{fedadam}, FedDisco \citep{feddisco}, FedLAW \cite{fedlaw} and L-DAWA \citep{ldawa} operate on the server side. Where FedDisco and L-DAWA specifically emphasize the aggregation scheme adjustments, making them highly relevant to our proposed method.
Additionally, we also show the performance of FedLAW \citep{fedlaw}. However, since it leverages additional data for fine-tuning that other methods do not, we present it only for reference. More details about the experimental setup can be found in Appendix \ref{exp_details}.\footnote{The source code is available at \url{https://github.com/ChanglongShi/FedAWA}}
To emulate the federated learning scenario, we randomly partition the training dataset into $K$ groups, assigning group $k$ to client $k$.
In practical FL scenarios, clients often exhibit heterogeneity, resulting in Non-IID characteristics in their data. To simulate this heterogeneity, we employ Dirichlet sampling, denoted as $Dir_{\alpha}$, which is widely used in FL literature \citep{fedma, dirc, feddisco}. A smaller $\alpha$ value corresponds to greater Non-IID characteristics. For a fair comparison, we apply the same data synthesis approach across all methods. 

\vspace{-0.5em}
\subsection{Main Results}\label{perfoeva}
\textbf{Performance.} In this section, we compare our proposed FedAWA with all baselines and report the test accuracy on all datasets, as shown in Table \ref{main_res}. 
It can be observed that FedAWA achieves the overall best performance across various datasets and heterogeneity settings, highlighting the effectiveness of our proposed method. 
In addition, FedAWA-L achieves better performance compared to FedAWA, indicating that the layer-wise approach enables finer model aggregation, which provides a more detailed and precise optimization process, leading to improved results. However, this improvement comes at the cost of higher computational overhead, as aggregation weights must be optimized separately for each layer of the model.
Therefore, FedAWA strikes a more balanced trade-off between performance and computational cost.
To evaluate the robustness of our method in different federated learning scenarios, we tune three crucial parameters of FL: the number of clients $K \in$ \{10, 30, 50\}, the number of local epoch $E \in$ \{1, 5, 10\}, and partial participation ratio $R \in$ \{0.1, 0.3, 0.5\}. We show the result in Figure \ref{fig_diff_fl_set}. The experiments consistently reveal that our proposed method consistently brings performance improvement across different FL settings.


\begin{figure}[t]
  \centering
   \includegraphics[width=0.46\textwidth]{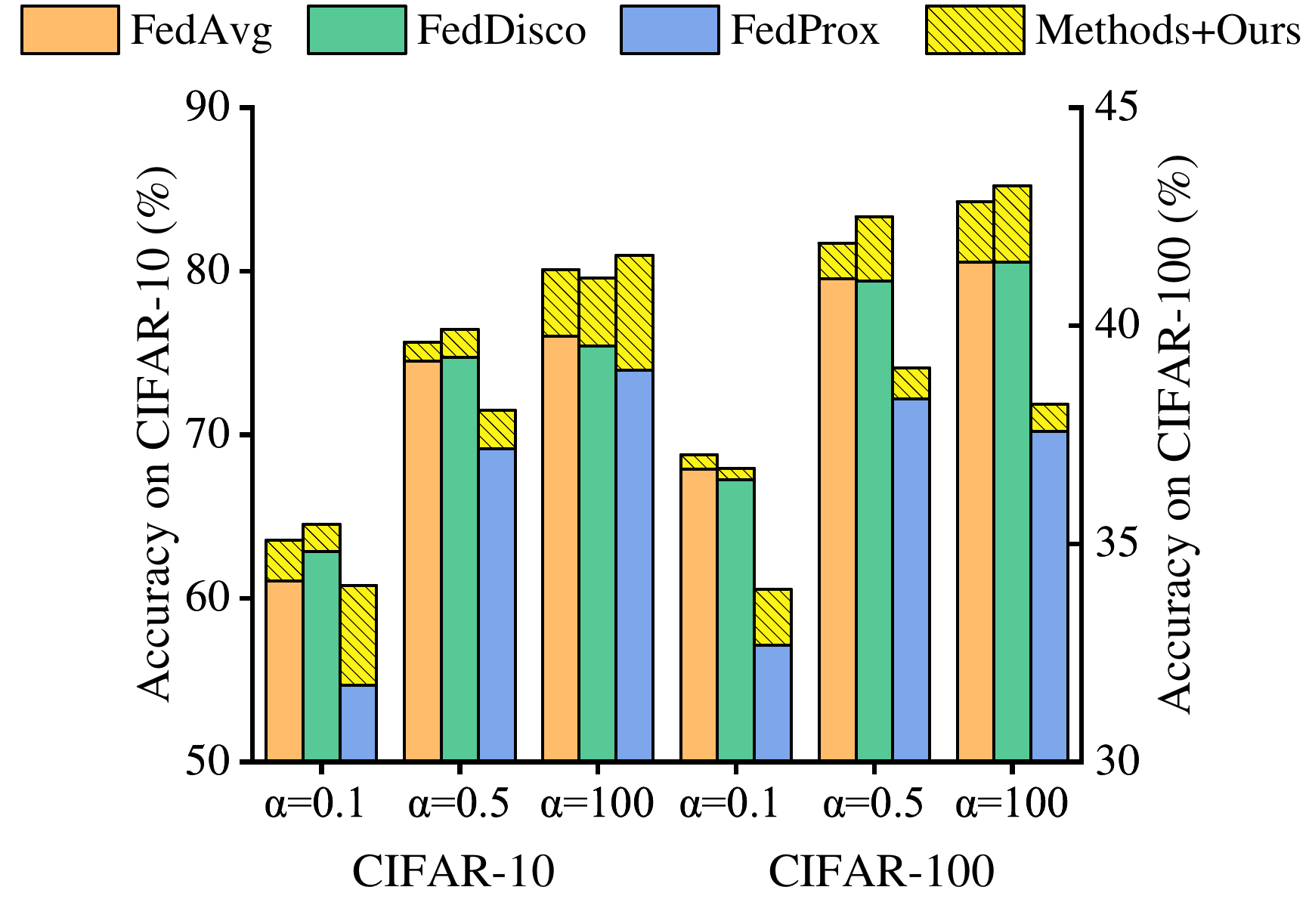}
    \vspace{-0.5em}
   \caption{Modularity. Performance improvements achieved by integrating our proposed FedAWA with different FL algorithms under varying datasets and degrees of data heterogeneity.
  }
  \vspace{-1em}
   \label{combine_res}
\end{figure}

\textbf{Modularity.} Our proposed FedAWA can be integrated with various existing FL methods to further enhance their performance. As illustrated in Figure \ref{combine_res}, we evaluated the performance of our approach in combination with the foundational FL algorithm FedAvg, the client-side adjustment method FedProx, and the aggregation weight adjustment method FedDisco. Specifically, when combined with FedDisco, FedAWA initializes the aggregation weights using FedDisco's approach and subsequently optimizes them further using our proposed method. The results demonstrate that FedAWA consistently improves the performance of the baseline methods across varying degrees of data heterogeneity, further validating the effectiveness of our approach.

\begin{table}[t]
  \begin{center}
    \caption{Average aggregation execution time. (ResNet20) 
    }
    \vspace{-1em}
    \label{time}
    \makebox[1\linewidth][c]{
    \resizebox*{1\linewidth }{!}{
    \renewcommand{\arraystretch}{1.5}
    \begin{tabular}{c c c c c c c c c c c c c c c c c c c c c c }
    \\
    \toprule
        Method&FedAvg \cite{fedavg}&L-DAWA \cite{ldawa}&FedLAW \cite{fedlaw}&\textbf{FedAWA}&\textbf{FedAWA-L}\\
      \midrule
        Execution Time (Sec)&0.10&2.52&10.11&0.82&15.21\\
    \bottomrule
    \end{tabular}
    }
    }
    \vspace{-1em}
  \end{center}
\end{table}

\textbf{Computation Efficiency.}
In Table \ref{time}, we show the aggregation execution time of our method FedAWA, FedAWA-L and the closely related work FedAvg \cite{fedavg}, L-DAWA \cite{ldawa}, and FedLAW \citep{fedlaw}. The used model architecture is ResNet20 and the dataset is CIFAR-10. For FedLAW, the proxy dataset contains 200 samples and the server epoch is 100 (consistent with \citep{fedlaw}).
It can be observed that, while FedAWA-L achieves the best performance, it incurs additional computational overhead compared to other methods. In comparison to L-DAWA and FedLAW, our proposed FedAWA significantly reduces execution time. While FedAWA incurs slightly higher computational costs than FedAvg, the balance between its computational requirements and the resulting performance remains acceptable.

\begin{table}[t]
  \begin{center}
    \caption{The performance of compared methods with different model architectures.}
    \vspace{-1em}
    \label{diff_backbone}
    \resizebox{\linewidth}{!}{
    \renewcommand{\arraystretch}{1}
    \begin{tabular}{l|c|ccccc}
   \toprule
       \textbf{Dataset}&\textbf{Method}&CNN&ResNet20&WRN56\_4&DenseNet121&ViT\\
      \midrule
         \multirow{5}{*}{\textbf{CIFAR-10}}&FedLAW \cite{fedlaw}&70.18&75.37&80.46&86.43&51.20\\
         
         &FedAvg \cite{fedavg}&68.28&75.07&78.97&86.14&51.31\\

         &FedDisco \cite{feddisco}&70.24&73.42&78.74&85.05&53.97\\
         
        &L-DAWA\cite{ldawa}&70.87&75.79&81.51&86.29&53.43\\

        &\textbf{FedAWA (Ours)}&\textbf{71.17}&\textbf{77.71}&\textbf{81.96}&\textbf{86.63}&{\textbf{56.16}}\\
        \midrule
        \multirow{5}{*}{\textbf{CIFAR-100}}&FedLAW \cite{fedlaw}&33.59&38.53&18.44&55.36&22.03\\
        &FedAvg \cite{fedavg}&32.53&41.18&39.71&56.59&25.60\\

        &FedDisco \cite{feddisco}&33.57&40.23&45.96&58.91&31.38\\
        
        &L-DAWA \cite{ldawa}&32.55&41.23&48.22&61.45&30.71\\

        
        &\textbf{FedAWA (Ours)}&\textbf{37.38}&\textbf{41.79}&\textbf{49.21}&\textbf{62.61}&{\textbf{31.44}}\\
    \bottomrule
\end{tabular}}
  \end{center}
  \vspace{-2em}
\end{table}

\vspace{-0.5em}
\subsection{Ablation Studies}

\textbf{Effects of Model Architectures.} 
In Table \ref{diff_backbone}, we evaluate our proposed FedAWA across a diverse range of model architectures, including CNN, ResNet \citep{resnet}, Wide-ResNet (WRN) \citep{wrn}, DenseNet \citep{densenet}, and Vision Transformer (ViT) \citep{vit}. 
The results highlight the effectiveness of FedAWA across these different architectures, demonstrating its robust performance not only as network depth and width increase but also when applied to models with distinct architectural designs.

 \begin{table}[t]
  \begin{center}
    \caption{Comparison of directly using cosine similarity as the aggregation weights.}
    \label{cos_as_weight}
    \vspace{-1em}
    \resizebox{\linewidth}{!}{
    \renewcommand{\arraystretch}{1}
   \begin{tabular}{lccccc}
    \toprule
       {\textbf{Dataset}}&\multicolumn{2}{c}{CIFAR-10}&\multicolumn{2}{c}{CIFAR-100}&\multirow{2}{*}{\textbf{Average}}\\
       \cmidrule(lr){2-3} \cmidrule(lr){4-5}
       {\textbf{Heterogeneity}}& \multicolumn{1}{c}{$\alpha$=100} & \multicolumn{1}{c}{$\alpha$=0.1}& \multicolumn{1}{c}{$\alpha$=100}& \multicolumn{1}{c}{$\alpha$=0.1}\\
       
      \midrule
     
        FedAvg \cite{fedavg}&76.01&61.04&41.46&36.71&53.81\\
        L-DAWA \cite{ldawa}&76.10&62.87&42.39&36.31&54.42\\
        FedAWA-COS&78.89&62.14&42.31&36.74&55.02\\
        \textbf{FedAWA (Ours)}&80.10&63.55&42.84&37.04&\textbf{55.88}\\
    \bottomrule
\end{tabular}}
  \end{center}
  \vspace{-1em}
\end{table}

\textbf{Effects of Optimization.} In this section, we conduct experiments to analyze the effectiveness of the optimization process. For comparison, we implement FedAWA-COS, where the aggregation weights are determined by directly computing the similarity between the client vectors and the global vector, rather than optimizing them using Equation \eqref{loss}. In Table \ref{cos_as_weight}, with FedAvg and L-DAWA as the baseline, we compare the performance of FedAWA-cos and FedAWA across different datasets and levels of data heterogeneity. The results in Table \ref{cos_as_weight} demonstrate that FedAWA achieves higher model accuracy than FedAWA-COS, highlighting the benefits of iteratively optimizing the model aggregation weights to further improve model performance. Furthermore, compared to L-DAWA, which uses the model parameter similarity as the aggregation weight, FedAWA-COS achieves better performance, highlighting the client vector's ability to capture local data more effectively.

\begin{figure}[t]
  \centering
   \vspace{-1em}
   \includegraphics[width=0.45\textwidth]{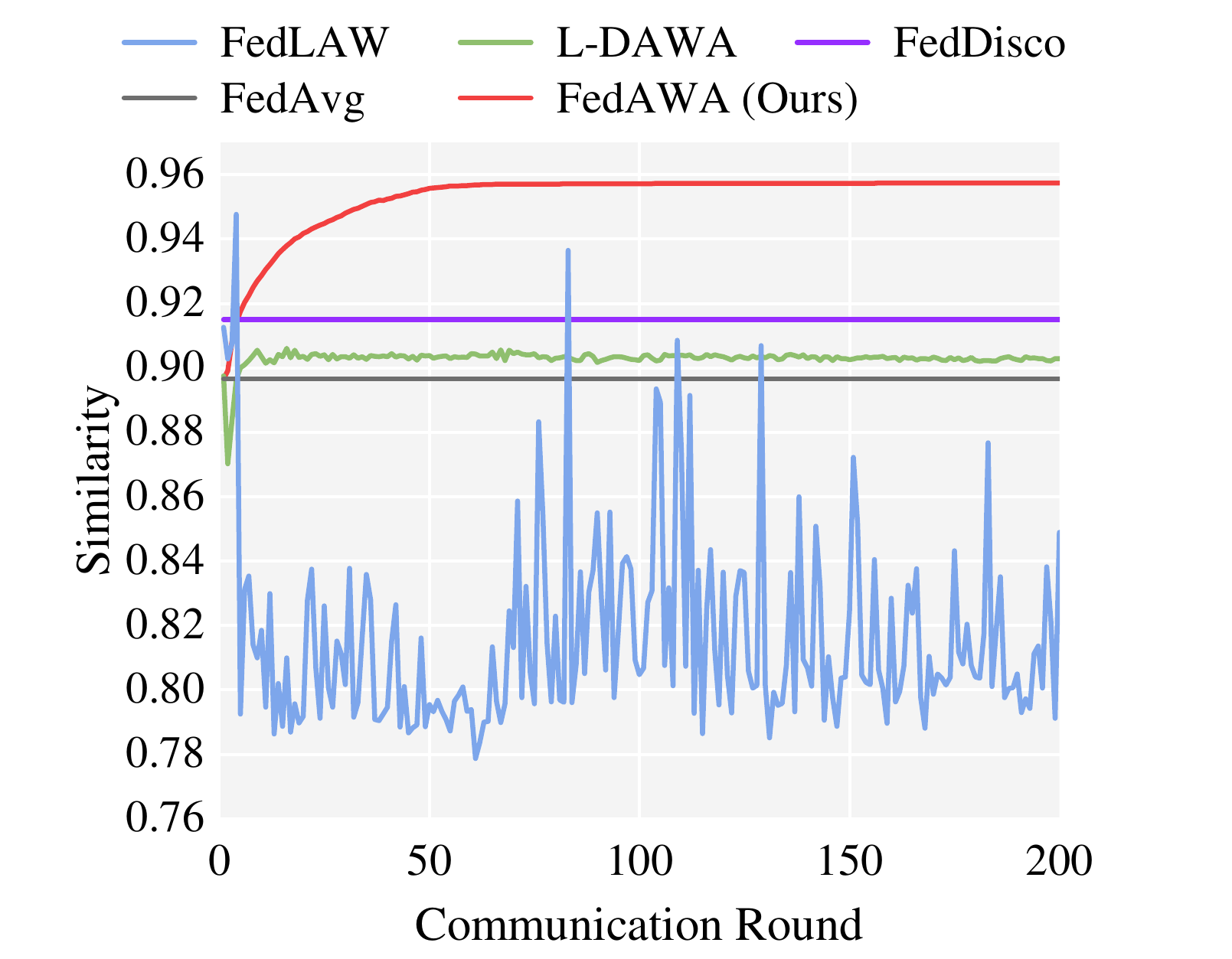}
    \vspace{-0.5em}
   \caption{Similarity between aggregation weights and dataset vectors during training.
  }
   \label{similarity}
   \vspace{-1.5em}
\end{figure}

\textbf{Aggregation Weights.} 
In this section, we conduct experiments to observe the evolution of aggregation weights during the training process. We first calculate the similarity between the client's local dataset and the global dataset, forming a K-dimensional dataset vector. We treat this dataset vector as the ideal aggregation weights vector, where local datasets more similar to the global dataset receive higher weights, and vice versa. We then observe the similarity between the aggregation weights obtained by different methods and the dataset vector. A higher similarity with the dataset vector indicates better alignment with the ideal aggregation weights. The results are shown in Figure \ref{similarity}. As can be seen, our method steadily increases and eventually converges during training, with a higher similarity to the dataset vector compared to other methods, demonstrating the effectiveness of our FedAWA. FedAvg and FedDisco use fixed aggregation weights, so there is no change in similarity during training. L-DAWA exhibits a sharp initial decline followed by an increase, which can be attributed to its direct use of model similarity as aggregation weights, introducing greater randomness during the early stages of training. FedLAW displays considerable fluctuations, likely due to the influence of the shrinkage factor within its optimization process.
More details can be found in Appendix \ref{app_aggre}.

 \begin{table}[t]
  \begin{center}
    \caption{The performance with varying distance metrics.}
    \vspace{-1em}
    \label{diff_distance}
    \resizebox{\linewidth}{!}{
    \renewcommand{\arraystretch}{1}
    \begin{tabular}{l|c|cccc}
   \toprule
       \textbf{Dataset}&\textbf{Method}&NIID($\alpha=0.1$)&NIID($\alpha=0.5$)\\
      \midrule
         
         \multirow{3}{*}{\textbf{CIFAR-10}}
         &FedAWA-w/o reg&63.05&75.06\\

         &FedAWA-w/ euc\_reg&63.53&75.35\\
       &FedAWA-w/ cos\_reg&\textbf{63.55}&\textbf{75.65}\\
        \midrule
        \multirow{3}{*}{\textbf{CIFAR-100}}&FedAWA-w/o reg&36.75&41.36\\

        &FedAWA-w/ euc\_reg&\textbf{37.20}&41.55\\
         &FedAWA-w/ cos\_reg&37.04&\textbf{41.89}\\
    \bottomrule
\end{tabular}}
  \end{center}
  \vspace{-2em}
\end{table}

\textbf{Effects of Regularization.} 
We investigate the impact of the regularization term (the second term) in Equation \ref{loss} on model performance. The results are presented in Table \ref{diff_distance}, where we show the result for three different methods: without regularization term, using the Euclidean distance for the regularization term, and using 1 - cosine similarity for the regularization term, labeled as FedAWA-w/o reg, FedAWA-w/ euc\_reg, and FedAWA-w/ cos\_reg, respectively. As observed, omitting the regularization term leads to a decrease in performance, which demonstrates the effectiveness of the regularization. Furthermore, FedAWA-w/ cos\_reg generally performs slightly better than FedAWA-w/ euc\_reg. This might be because the 1-cosine similarity ranges from 0 to 2, providing more stability compared to the unbounded range of Euclidean distance.

\vspace{-0.5em}
\section{Conclusion} \label{conclusion}

In this paper, through empirical explorations, we demonstrate that client vectors in federated learning effectively capture relevant information about local datasets. Furthermore, we investigate the relationship between the ideal model update direction and the client vector. Building on these observations, we propose FedAWA, a method that adaptively optimizes aggregation weights without relying on a proxy dataset, thereby enhancing model performance while addressing privacy concerns. Experimental results show that FedAWA delivers outstanding performance across various settings. A potential limitation of our method lies in its applicability exclusively to scenarios where client model architectures are identical. Model heterogeneity among clients remains a prominent challenge in federated learning, and this limitation is shared by many existing FL methods. As part of future work, we aim to extend our approach to accommodate scenarios with heterogeneous client model architectures.

\clearpage
{
\section*{Acknowledgements}
We truly thank the reviewers for their great effort in our submission. Changlong Shi, Bingjie Zhang, Dandan Guo and Yi Chang are supported by the National Natural Science Foundation of China (No. U2341229, No. 62306125) and the National Key R\&D Program of China under Grant (No. 2023YFF0905400).

    \small
    \nocite{li2024uncertaintyrag}
    \bibliographystyle{ieeenat_fullname}
    \bibliography{reference}
}

\clearpage
\setcounter{page}{1}
\maketitlesupplementary
\renewcommand{\thesection}{\Alph{section}}  

\begin{figure*}[htb]
    \centering 
    \begin{subfigure}[b]{0.35\linewidth}
        \includegraphics[width=\linewidth]{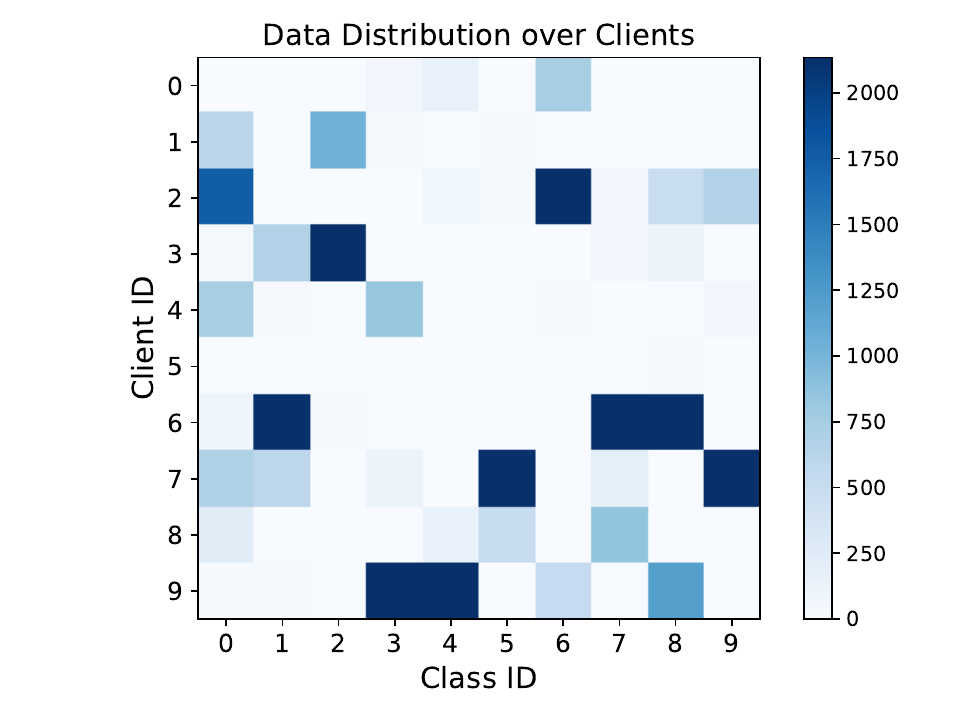}
        \caption{FashionMNIST, $\alpha=0.1$.}
    \end{subfigure}%
    \begin{subfigure}[b]{0.35\linewidth}
        \includegraphics[width=\linewidth]{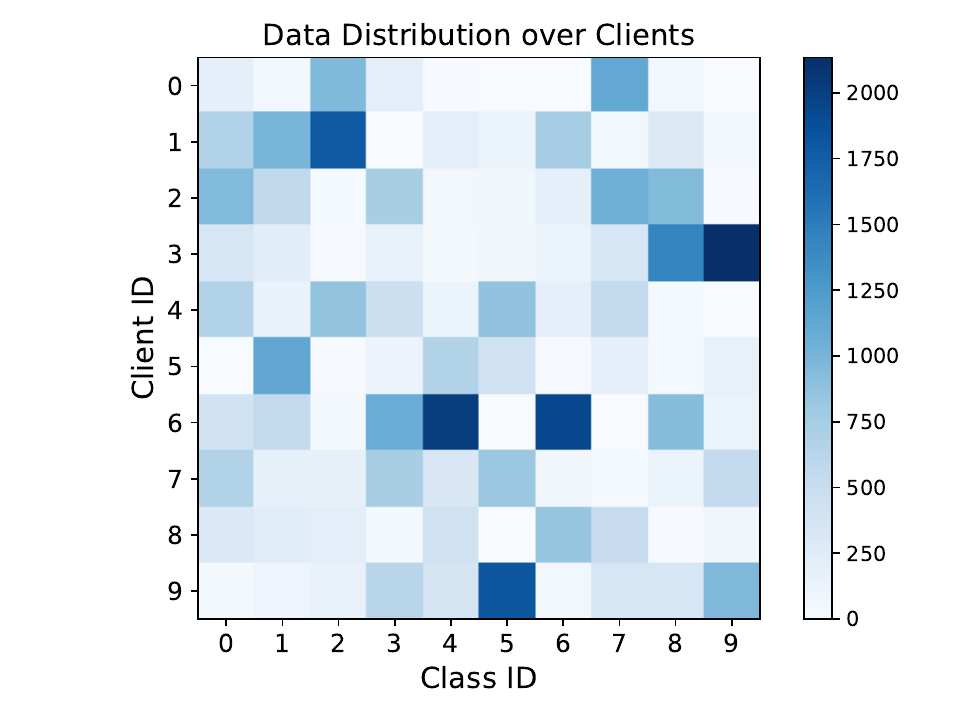}
        \caption{FashionMNIST, $\alpha=0.5$.}
    \end{subfigure}%
    \begin{subfigure}[b]{0.35\linewidth}
        \includegraphics[width=\linewidth]{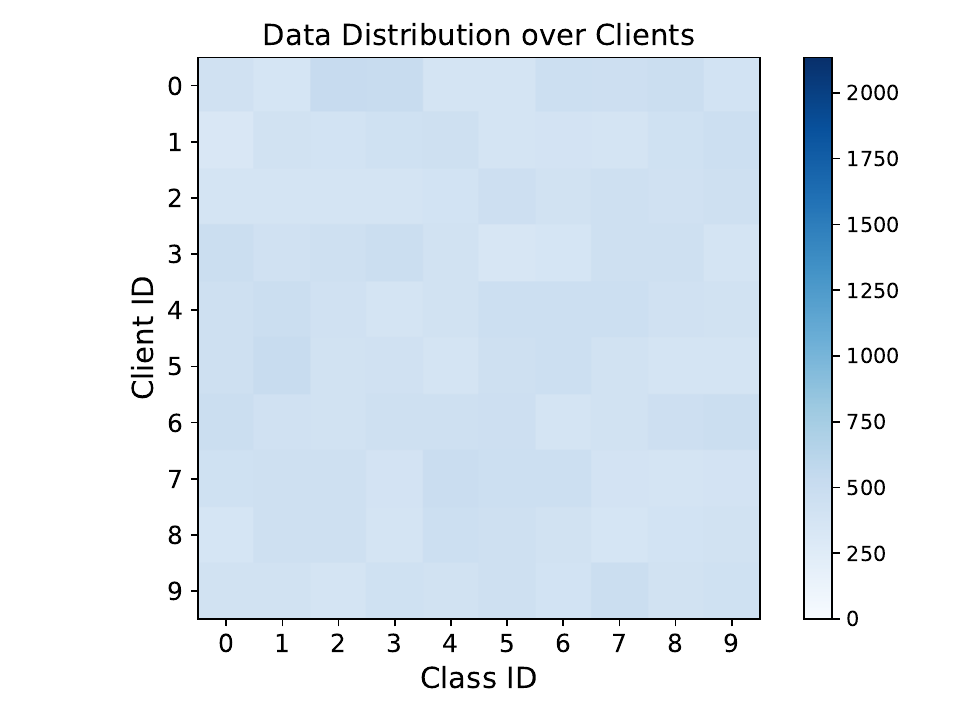}
        \caption{FashionMNIST, $\alpha=100$.}
    \end{subfigure}
    \vskip 1em
    \begin{subfigure}[b]{0.35\linewidth}
        \includegraphics[width=\linewidth]{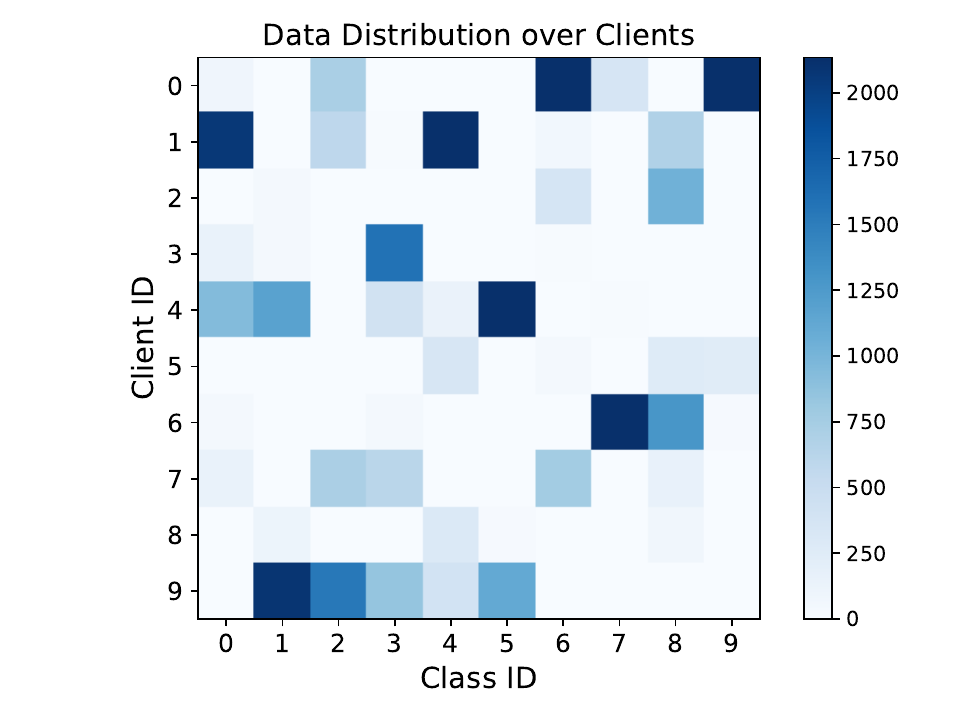}
        \caption{CIFAR-10, $\alpha=0.1$.}
    \end{subfigure}%
    \begin{subfigure}[b]{0.35\linewidth}
        \includegraphics[width=\linewidth]{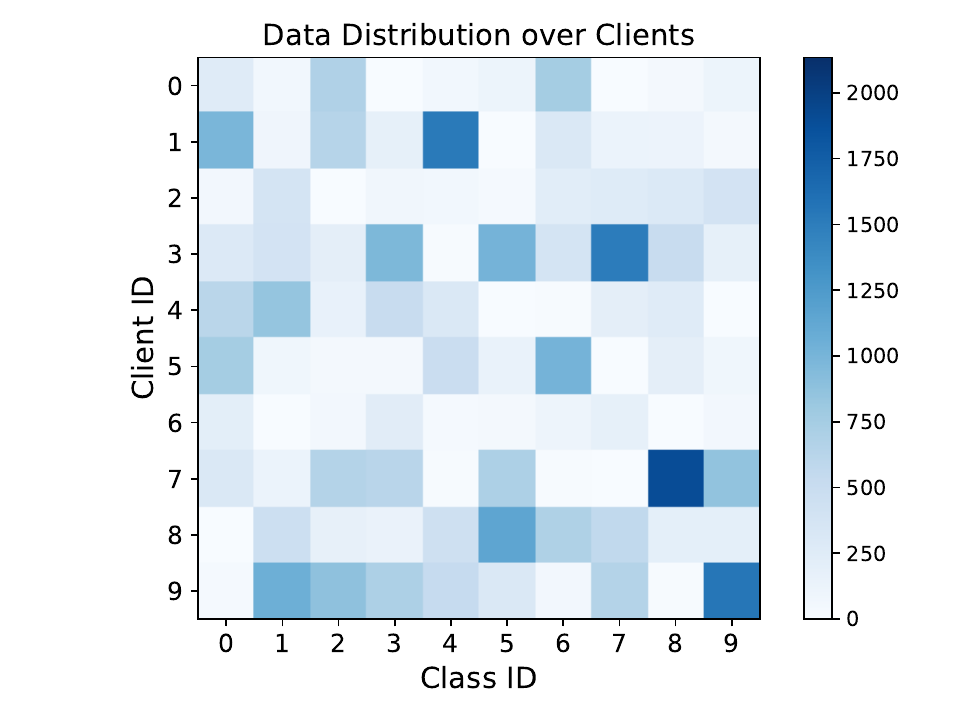}
        \caption{CIFAR-10, $\alpha=0.5$.}
    \end{subfigure}%
    \begin{subfigure}[b]{0.35\linewidth}
        \includegraphics[width=\linewidth]{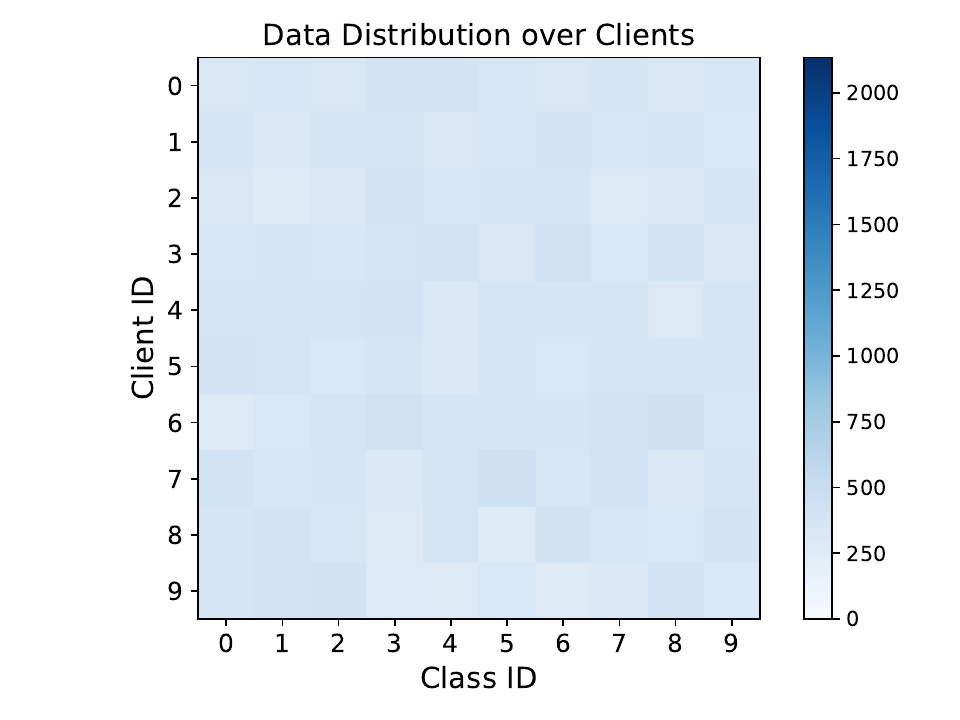}
        \caption{CIFAR-10, $\alpha=100$.}
    \end{subfigure}
    \caption{Data distribution over categories and clients.}
    \vspace{-1em}
    \label{datadis}
\end{figure*}
\setcounter{section}{0}
\section{Experiment details}\label{exp_details}
In this section, we provide the details of the experimental setup, environment, datasets, and model architectures used in this paper.
\subsection{Client Vector and Local Data.}\label{app_client_vector}
In Section \ref{observation}, we demonstrated experimentally the relationship between the client vector and local data. In this section, we will provide a more detailed explanation of the experimental setup and offer a further analysis of the results. The Experiments were conducted to verify whether the client vector reflects information about the local data.
In the experiment, we set up 12 clients, and the dataset we used was CIFAR-10. To observe the data differences more intuitively, we introduced an extreme scenario of data heterogeneity: The local data of clients 1 to 4 contained only the first 5 classes of the CIFAR-10, clients 5 to 8 had only the left 5 classes, and clients 9 to 12 had local datasets that included all classes. The size of the local dataset for each client was the same.
We calculated the distances between the client local datasets via optimal transport \citep{ot}, and the results are displayed in Figure \ref{matrix_data}.

Then, we conducted federated learning training, during which each client obtained its own client vector $\tau_k$ after local training. We then compared the distance between the client vectors, with the results displayed in Figure \ref{matrix_client_vector}. The relationships between the client vectors closely resemble those of the local data distributions. For example, client 1’s client vector exhibits minimal differences with clients 2-4 due to their similar local data distributions. However, the differences between client 1 and clients 5-8 are much larger because of their highly divergent data distributions: client 1’s local data contains only the first 5 classes, while clients 5-8 have only the last 5 classes.  The differences between client 1 and clients 9-12 are smaller than those with clients 5-8, as clients 9-12 include data from all classes, making their distribution relatively closer to client 1. This demonstrates that the client vector can effectively capture relevant information about the local data. Hence, we explored the possibility of leveraging this phenomenon to enhance the model aggregation process in federated learning.

If the distance is computed directly using the overall model parameters, the results, as shown in Figure \ref{matrix_model}, indicate that the distances between models are relatively similar. This is because each client model is optimized from the same global model, and the parameter variations are small relative to the overall model parameters. As a result, the local models do not exhibit significant differences after training, making it difficult to effectively capture the relationships among the local datasets. It is important to note that for both Figure \ref{matrix_client_vector} and Figure \ref{matrix_model}, we compute the distance between clients vectors and model parameters using 1 - cosine similarity. This method normalizes the values to a consistent scale (ranging from 0 to 2), allowing for a more intuitive comparison of the differences between the models.

\subsection{Aggregation Weights.}\label{app_aggre}
In this section, we provide more details regarding the experiment in Figure \ref{similarity}. 
We first calculate the similarity between the local dataset and the global dataset, where the partitioning of the local dataset is the same as in Appendix \ref{app_client_vector}, and the global dataset is the union of all local datasets. We use a pre-trained ResNet20 to extract features from the datasets and compute the distance between the two datasets using Optimal Transport \cite{ot}. Since our goal is to measure the similarity between datasets, we convert the OT distance into a similarity score as:
\begin{equation}
\begin{aligned}
\label{ot_sim}
\text{Similarity}(P, Q) = \frac{1}{1 + d_{OT}(P, Q)},
\end{aligned}
\end{equation}
where $P$ and $Q$ represent the distributions of local data and global data, respectively, while $d_{OT}(\cdot,\cdot)$ denotes the optimal transport distance.
This results in a k-dimensional dataset vector representing the similarity between each local dataset k and the global dataset. This vector depends solely on the datasets and remains fixed throughout training. The k-th element in the vector indicates the similarity between the local dataset k and the global dataset. We use this as the ideal aggregation weights, assigning higher aggregation weights to datasets more similar to the global dataset, and vice versa \cite{feddisco}. We then evaluate the aggregation weights of different methods by calculating the cosine similarity between the aggregation weights and the data vector. As shown in Figure \ref{similarity}, the results demonstrate the effectiveness of our method.

\subsection{Datasets}

In the experiment, we utilized four image classification datasets: CIFAR-10 \citep{cifar}, CIFAR-100 \citep{cifar}, and Tiny-ImageNet \citep{tinyimagenet}, which have been widely employed in prior Federated Learning methods \citep{fedlaw, feddisco, ccvr}. All these datasets are readily available for download online.
To generate a non-IID data partition among clients, we employed Dirichlet distribution sampling $Dir_{\alpha}$ in the training set of each dataset, the smaller the value of $\alpha$, the greater the non-IID. In our implementation, apart from clients having different class distributions, clients also have different dataset sizes, which we believe reflects a more realistic partition in practical scenarios. We set $\alpha=$0.1, 0.5, and 100, respectively.
When $\alpha$ is set to 100, we consider the data to be distributed in an IID manner.
The data distribution across categories and clients is illustrated in Figure \ref{datadis}. Due to the large number of categories, we did not display the data distribution of CIFAR-100 and Tiny-ImageNet. Their distributions are similar to the other two datasets.

\begin{table}[!t]
    \centering
    \caption{Long-tail Scenarios.}
    \label{longtail}
        \renewcommand{\arraystretch}{1}
        \begin{tabular}{lcc}
            \toprule
            \textbf{Method} & \textbf{CIFAR10-LT} & \textbf{CIFAR100-LT} \\
            \midrule
            FedAvg         & 77.45 & 45.87 \\
            CReFF          & 80.71 & 47.08 \\
            CLIP2FL        & 81.18 & 48.20 \\
            \rowcolor{Gray} 
            \textbf{CReFF+AWA(Ours)} & \textbf{83.11} & \textbf{49.63} \\
            \bottomrule
        \end{tabular}
  
\end{table}

\begin{table}[!t]
    \centering
    \caption{Multi-domain Scenarios.}
    \label{multi_domain}
    \resizebox{\linewidth}{!}{
        \renewcommand{\arraystretch}{0.7}
        \begin{tabular}{lcccc|c}
            \toprule
            \textbf{Methods} & \textbf{SVHN} & \textbf{USPS} & \textbf{MNIST} & \textbf{SYN} & \textbf{AVG}  \\
            \midrule
            FedAvg    & 76.56 & 90.85 & 98.14 & 55.01 & 80.14 \\
            FedProx   & 77.01 & 90.24 & 98.11 & 56.66 & 80.50 \\
            FedProto  & 80.35 & 92.44 & 98.30 & 53.58 & 81.16 \\
            FPL       & 80.27 & 92.71 & 98.31 & 61.20 & 83.12 \\
            \rowcolor{Gray} 
            \textbf{FPL+AWA(Ours)} & 80.63 & 91.58 & 97.76 & 70.40 & \textbf{85.09}\\
            \bottomrule
        \end{tabular}}

\end{table}

\begin{table}[!t]
    \centering
    \caption{Text Classification.}
    \label{text_cls}
        \renewcommand{\arraystretch}{0.85}
        \begin{tabular}{cccccc}
            \toprule
            \multirow{2}{*}{\textbf{Method}} & \multicolumn{2}{c}{\textbf{AG News}} & \multicolumn{2}{c}{\textbf{Sogou News}} \\
            \cmidrule(lr){2-3} \cmidrule(lr){4-5}
            & $\alpha = 0.1$ & $\alpha = 0.5$ & $\alpha = 0.1$ & $\alpha = 0.5$  \\
            \midrule
            FedAvg   & 73.43 & 70.37 & 87.68 & 91.53 \\
            FedProx  & 65.07 & 74.56 & 88.60 & 92.28 \\
            \rowcolor{Gray} 
            \textbf{FedAWA}   & \textbf{77.25} & \textbf{80.23} & \textbf{90.85} & \textbf{94.09} \\
            \bottomrule
        \end{tabular}
\end{table}

\subsection{Experiment}

In this section, we conducted additional experiments across long-tail, multi-domain, and text classification scenarios to further evaluate the performance of the proposed model under more scenarios.

To further evaluate the performance of the proposed model under more complex data heterogeneity scenarios, we performed comparisons and integrations with algorithms specifically designed for these challenging scenarios. These experiments focused on two primary settings: global long-tail data distribution and multi-domain data distribution.

For the long-tail data distribution scenario, experimental results are presented in Table \ref{longtail}. In these experiments, the proposed algorithm FedAWA was combined with the CReFF \cite{CReFF} algorithm and compared with federated learning methods designed to address long-tail data distributions, such as CReFF \cite{CReFF} and CLIP2FL \cite{clip2fl}.

Similarly, for the multi-domain data distribution scenario, the results are shown in Table \ref{multi_domain}. Here, FedAWA was integrated with the FPL \cite{FPL} algorithm and compared with federated learning methods specifically tailored for multi-domain distributions, namely FPL \cite{FPL} and FedProto \cite{fedproto}.

The experimental results demonstrate that FedAWA consistently improves model performance even in more complex heterogeneous data environments, thereby confirming the stability and robustness of the proposed method.

To further demonstrate the applicability of our method to textual modalities, we conducted additional experiments on NLP datasets AG News \cite{agnews} and Sogou News \cite{agnews} under various data heterogeneity settings. As shown in Table \ref{text_cls}, FedAWA consistently outperforms baseline methods in text classification tasks.

\subsection{Hyperparameters}
If not mentioned otherwise, The number of clients, participation ratio, and local epoch are set to 20, 1, and 1, respectively. 
We set the initial learning rates as 0.08 and set a decaying LR scheduler in all experiments; that is, in each round, the local learning rate is 0.99*(the learning rate of the last round).
We adopt local weight decay in all experiments. We set the weight decay factor as 5e-4.
We use SGD optimizer as the clients’ local optimizer and set momentum as 0.9.

\subsection{Models}
For each dataset, all methods are evaluated with the same model architectures for a fair comparison.
In Table \ref{main_res}, We use ResNet20 \citep{resnet} for CIFAR-10 and CIFAR-100, ResNet18 for Tiny-ImageNet.
In Table \ref{diff_backbone}, we compare the experimental results of different model architectures. The specific model architectures are as follows:
  
\textbf{CNN.} The CNN is a convolution neural network model with ReLU activations. In this paper CNN consists of 3 convolutional layers followed by 2 fully connected layers. The first convolutional layer is of size (3, 32, 3) followed by a max pooling layer of size (2, 2). The second and third convolutional layers are of sizes (32, 64, 3) and (64, 64, 3), respectively. The last two connected layers are of sizes (64*4*4, 64) and (64, num\_classes), respectively.

\textbf{ResNet, WRN, DenseNet and ViT.} We followed the model architectures used in \citep{fedlaw,modelarc,vit}. The numbers of the model names mean the number of layers of the models. Naturally, the larger number indicates a deeper network. For the Wide-ResNet56-4 (WRN56\_4) in Table \ref{diff_backbone}, "4" refers to four times as many filters per layer.

\balance



\end{document}